\documentclass[runningheads]{llncs}
\usepackage{booktabs} 
\usepackage{amsmath}
\usepackage{amsfonts}
\usepackage{graphicx}
\usepackage{color}
\usepackage{hyperref}
\usepackage[belowskip=-2pt,skip=0.1pt]{caption}
\usepackage[caption=false]{subfig}
\usepackage{algorithm}
\usepackage{algpseudocode}
\usepackage{balance}
\usepackage[numbers]{natbib}

\usepackage{tikz} 
\usepackage{xcolor}

\DeclareMathOperator*{\argmax}{argmax}
\DeclareMathOperator*{\argmin}{argmin}

\begin{document}
\title{Discovering patterns of online popularity from time series}
%
%
\author{Mert Ozer\inst{1}\orcidID{1234-5678-9012} \and
Anna Sapienza\inst{2} \and
Andr\'{e}s Abeliuk\inst{2} \and
Goran Muric \inst{2} \and
Emilio Ferrara \inst{2}}
\authorrunning{M. Ozer et al.}
%
\institute{School of Computing, Informatics, and Decision Science Engineering, Arizona State University, Tempe, USA
\\\email{mozer@asu.edu}\\ \and
Information Sciences Institute, University of Southern California, Marina del Rey, USA\\
\email{\{annas,gmuric,aabeliuk,ferrarae\}@isi.edu}}
\maketitle              
\begin{abstract}
How is popularity gained online? Is being successful strictly related to rapidly becoming viral in an online platform or is it possible to acquire popularity in a steady and disciplined fashion? What are other temporal characteristics that can unveil popularity of online content? To answer these questions, we leverage a multi-faceted temporal analysis of the evolution of popular online contents. Here, we present dipm-SC: a multi-dimensional shape-based time-series clustering algorithm with a heuristic to find the optimal number of clusters. First, we validate the accuracy of our algorithm on synthetic datasets generated from benchmark time series models. Second, we show that dipm-SC can uncover meaningful clusters of popularity behaviors in a real-world Twitter dataset. By clustering the multidimensional time-series of popularity of contents coupled with other domain-specific dimensions, we uncover two main patterns of popularity: bursty and steady temporal behaviors. Moreover, we find that the way popularity is gained over time has no significant impact on the final cumulative popularity.

\end{abstract}

\section{Introduction}
How do contents get popular on the Web? Is being successful, strictly constrained to a rapid and bursty virality~\cite{adar2004implicit} or is it possible to acquire popularity in a steady and disciplined fashion? What are other temporal characteristics that can unveil popularity of online contents?

Temporal dynamics of popularity have been studied extensively in Twitter~\cite{wu2007novelty}, news~\cite{lerman2010information}, and videos~\cite{crane2008robust}. Thus far, previous work studying popularity  either explored topological patterns in the networks that lead to cascades~\cite{bakshy2011everyone,leskovec2007patterns}, or focused on the short time period in which significant popularity is gained~\cite{ksc,lehmann2012dynamical}. 
However, popularity is subject to endogenous and exogenous factors~\cite{lehmann2012dynamical} that affects its temporal dynamics.
For instance, popularity of an item may depend on the amount of external promotions it receives~\cite{rizoiu2017expecting}.
Analyzing different temporal patterns that arise when considering both endogenous and exogenous factors will further extend our understanding on temporal collective behavior. 

To take into account all these factors at a time and tackle the problem of uncovering their relation with the dynamics of popularity, we focus on multidimensional clustering methods. Multidimensional clustering can indeed allow us to uncover the hidden relations between time series of events having different evolutions in more than one dimension. For instance, the dynamics of a user content can be not only described by its popularity but also by the online activity and pace of the user, number of social connections in social networks.

A wide range of studies have been focused on solving the problem of efficiently clustering time series~\cite{Aghabozorgi2015TimeseriesC}. Some approaches have adopted methods such as sub-sequence techniques~\cite{ushapelet,shapelet,ushapelet2}, and dimensionality reduction models~\cite{pca_based,locally_adaptive}. However, we aim at uncovering different patterns of popularity on the basis of both their shapes and the time at which the main popularity gain is achieved. Therefore, sub-sequence or subspace solutions are not suitable for this analysis as they lose temporal granularity and order. 
In this paper, we take inspiration from unidimensional clustering approaches~\cite{ksc,kshape,kdba,kavged} that use a shape-based distance and a time series similarity measure that is scale and shift invariant. We generalize these methods to multidimensional time series and parametrized the temporal shifting, thus allowing us to capture temporal patterns occurring at different times.

In the present work, we design a multi-faceted temporal analysis to study the timeline of popular online content and provide insights on how popularity is gained over time. To this aim, we extend K-Spectral Centroid (kSC)~\cite{ksc}, a unidimensional shape-based time series clustering algorithm into our multidimensional version, called dipm-SC. It partitions multidimensional time series data into clusters and uses a heuristic to find the optimal number of clusters. Each cluster is characterized by a centroid, having a unique shape per dimension identifying the average temporal behavior of the cluster members.

We validate the accuracy of our algorithm on the datasets generated from benchmark time series models with varying parameters of time series length, dimensions and number of underlying clusters. To the best of our knowledge, this is the first study which explores multidimensional use of k-SC.

Then, with the help of the dipm-SC, we study temporal multidimensional behavior in two real world large datasets containing data from GitHub and Twitter. In Github, we study the popularity of repositories considering a series of other temporal dimensions that keep track of the repository  development, such as pushes and pull requests a repository receives, and repository issues, such as issue entries and comments under them. Our algorithm identifies 9 main clusters among most popular repositories on GitHub in different shapes. In Twitter, we study how popular hashtags unfold in a window of time where a major event happens. We keep track of number of times a hashtag used hourly (as a measure of popularity) alongside with hourly positive and negative sentiment rates and tweets containing the hashtag posted by bot accounts. Our algorithm identifies 6 main clusters among most popular hashtags on Twitter. We show that both bursty and steady clusters are evident in both datasets. On Twitter, when a bursty popularity gain is present, we observe increased sentiment and bot activity rates around the popularity spike. We also note that sentiment and bot activity decay after the spike can be characterized by the fat-tail of the popularity spike. Moreover, we find that how repositories unfold over time shows no significant impact on the cumulative popularity at the end of their lifetime. However, in the Twitter data, our algorithm is able identify different cluster shapes leading to different cumulative popularity gains.

\paragraph{Our contributions} 
\begin{enumerate}
\item We propose a novel algorithm to cluster multidimensional shape-based time series. Our approach detect the interplay of different dimensions that allow to perform a non-supervised classification of different temporal patterns.
\item We validate the accuracy of our algorithm on the datasets generated from benchmark time series models, and compare its performance with other scalable multidimensional time series methods. 
\item We present two real-world case studies, where we showcase the effectiveness of our algorithm in both finding meaningful popularity dynamics and providing clusters that are coherent and easy to interpret.
\end{enumerate}
\vspace{-0.2cm}

\section{Methodology}
\label{met}
In this section, we formally define the problem and discuss the need for extensions to time series clustering algorithms in the literature. Then, we develop our multidimensional extension based on a several unidimensional shape-based time series clustering methods.

\subsection{Problem Definition}
In this work, we aim at finding temporal patterns of popular online content in a multi-faceted manner. Let $N$ be the number of entities to be studied described by $D$ temporal features. Each temporal feature is represented by a discrete time series in $M$ time units (e.g., days or hours). This representation of our data can be encoded in a tensor $\mathcal{X}$ of size $N\times D\times M$. 

Given $\mathcal{X}$, we want to partition the $N$ multivariate time series into an optimal number of $K$ clusters, by
taking the temporal dynamics of different $D$ dimensions into account at the same time. To this aim, we perform a
clustering process based on the cumulative similarity of shapes among corresponding dimensions of the time series.
Therefore, each cluster $k$ is represented by a multi-dimensional shape $\mathbf{c}_{k}\in \mathbb{R}^{D\times M}$,
which describes the overall behavior of the cluster across multiple dimensions. In this work, solving the
aforementioned problem gives us the ability to uncover similar patterns of popularity gain while differing in other
dimensions, and vice versa.

\subsection{Proposed Algorithm}

The kSC algorithm proposed by~\citet{ksc} stands out as a prominent solution for analyzing time series data constructed from online human behaviors. Along that line, there have been significant contributions to the shape-based time series clustering literature~\cite{kshape,kms}. Here, we extend and compare the following shape-based univariate time series clustering algorithms: kShape\cite{kshape}, kDBA\cite{kdba}, kAVG+ED\cite{kavged}, and kSC\cite{ksc}. We only report the details about the methodology used to extend kSC\cite{ksc} as it is the only algorithm specifically developed for online social platforms and, as shown in the supplementary material \footnote{\url{www.public.asu.edu/~mozer/dipm-SC}}, is the one performing better in our synthetic scenario. However, we applied similar extensions to the kShape, kDBA, and kAVG+ED algorithms.

Our extension focuses on two parts. First, we modify the existing kSC algorithm to be applicable to multidimensional time series data. This process consists of the two steps of updating the distance and averaging functions of kSC. Second, we develop an iterative framework to find the optimal number of multidimensional shape clusters inspired by a similar work developed for kMeans~\cite{dipmeans}.

\subsubsection{kSC framework} K-Spectral Centroid (kSC) is an iterative kMeans based time series clustering algorithm. The algorithm iteratively assigns time series data to clusters and update cluster centroids until no updates to clusters is made. It adopts a scale and shift invariant distance function~\cite{ksc_distance} to measure the proximity of two time series. Therefore, each cluster's centroid represents the average shape of its members. At each iteration, the method uses a spectral clustering technique as its averaging function to update the centroids. However, as it is originally proposed and designed for unidimensional time series data, we need to extend its distance and averaging functions in order to apply it to a multidimensional scenario.


\subsubsection{Extending the Distance Function}
There are several ways to extend the univariate distance function to its multivariate version. First, we can compute distances separately on each dimension and sum them. However, this option is not optimal, as usual time series distances involve the manipulation of the signal through shifting and warping process. Thus, the use of separate distances would lead to different manipulations (shifting and warping) in each dimension. As our task requires to observe their interplay instead, manipulating each dimension differently is not desirable. Second, we can compute distances by using an optimal shared manipulation across all dimensions. Here, we propose extending the kSC algorithm's distance function~\cite{ksc_distance} as follows
\begin{equation}
dist(\mathbf{c}_k,\mathbf{x}) = \underset{\alpha_{d},q}{\min} \frac{1}{D}\sum_{d=1}^{D}\dfrac{\left\Vert\mathbf{c}_k(d,:) - \alpha_{d} \mathbf{x}_{q}(d,:) \right\Vert }{\left\Vert \mathbf{c}_k(d,:) \right\Vert } \;\;,
\label{eq:distance}
\end{equation}
where $\mathbf{x}(d,:)$ is the $d$'th dimension of time series $\mathbf{x} \in\mathbb{R}^{D\times M}$, $\mathbf{x}_{q}$ is the shifting of $\mathbf{x}$ by $q$ steps in time, and $\mathbf{c}_k$ is the centroid of cluster $k$.
The above equation constitutes the essential part of calculating the distance of multivariate time series $\mathbf{x}$ to a centroid. It indeed enforces the same shifting parameter $q$ regardless of dimensions. Yet, it keeps the scaling parameter $\alpha_d$ unique for each dimension. Here, we assume that even if each dimension may experience trends in different scales, they should have the same temporal order. 

Notice that, for any fix $q$, the optimal  $\alpha_d(q)$ that solves Eq.~\eqref{eq:distance} is
\begin{equation}
\alpha_d(q) = \frac{\mathbf{x}_{q}(d,:)^T\mathbf{c}_k(d,:)}{\left\Vert \mathbf{x}_{q}(d,:) \right\Vert^2}\;\;,
\label{eq:alpha}
\end{equation}  
by setting the gradient to zero while q is fixed. Thus, minimizing the distance function is reduced to a brute-force searching of the optimal $q$, where $\alpha_d$ is calculated for each shifting option as in Eq.~\eqref{eq:alpha}. The algorithm can become really slow when full-length shifting is allowed. As we are interested in temporal behavior of online content, we focus on moments in time when the events happen. Therefore, we aim to keep the shifting parameter to the minimum.
\subsubsection{Extending the Averaging Function}

As a second step in our method, we need to extend the averaging function of kSC. The role of the averaging function is to find a suitable centroid for each cluster, where the distance of its members is minimum to the centroid

\begin{align*}
\mathbf{c}_k = \underset{\mathbf{c}}{\argmin}\sum_{\mathbf{x} \in S_k}dist(\mathbf{c},\mathbf{x})^2\;\;,
\end{align*}
where $S_k$ is the set of time series belonging to cluster $k$. 

A possible way of extending the averaging function consists of finding a univariate centroid which minimizes the distance to every dimension. 
We avoid this option since there may be different behaviors on different dimensions and our task involves observing them in parallel. Instead, we extract a unique shape for each dimension separately. First, we replace the distance function with its multivariate version

\begin{align}
\mathbf{c}_k = \underset{\mathbf{c}}{\argmin}\sum_{\mathbf{x} \in S_k}\underset{\alpha_{d},q}{\min} \frac{1}{D^2}\sum_{d}^{D}\dfrac{\left\Vert \mathbf{c}(d,:) - \alpha_{d} \mathbf{x}_{q}(d,:)\right\Vert^2 }{\left\Vert \mathbf{c}(d,:) \right\Vert^2 } \;\;.
\label{eq:centroid}
\end{align}

To be able to use the spectral properties of Rayleigh quotient in accordance with the original kSC paper\cite{ksc}, we first need to find the optimal $\alpha_d$ and $q$. $\alpha_d$ has a direct solution as proposed in Eq.~\eqref{eq:alpha}. However $q$ is dimension invariant. Here we set $q$ as the optimal $q$ found during the distance's computation, and jointly shift every dimension of the time series $\mathbf{x}$ accordingly during the new centroid calculation. Finally, we invert the order of the two sums in Eq.~\eqref{eq:centroid} and get

\begin{align}
\mathbf{c}_k = \underset{\mathbf{c}}{\argmin}\frac{1}{D^2}\sum_{d}^{D}\sum_{\mathbf{x} \in S_k}\dfrac{\left\Vert \mathbf{c}(d,:) - \alpha_{d} \mathbf{x}(d,:)\right\Vert^2 }{\left\Vert \mathbf{c}(d,:) \right\Vert^2 } \;\;.
\end{align}

Our minimization problem does not have dimension independent variables anymore and we can solve each dimension's centroid separately by following similar steps to the univariate case

\begin{align*}
\mathbf{c}_k(d,:) &= \underset{\mathbf{c}'}{\argmin}\sum_{\mathbf{x} \in S_k}\dfrac{\left\Vert \alpha_{d} \mathbf{x}(d,:) - \mathbf{c}'\right\Vert^2 }{\left\Vert \mathbf{c}' \right\Vert^2 } \\
  &= \underset{\mathbf{c}'}{\argmin}\frac{1}{\left\Vert \mathbf{c}' \right\Vert^2 }\sum_{\mathbf{x} \in S_k}\left\Vert \dfrac{\mathbf{x}(d,:)^{T}\mathbf{c}'\mathbf{x}(d,:)}{\left\Vert\mathbf{x}(d,:)\right\Vert^2} - \mathbf{c}'\right\Vert^2 \\
    &= \underset{\mathbf{c}'}{\argmin}\frac{1}{\left\Vert \mathbf{c}' \right\Vert^2 }\sum_{\mathbf{x} \in S_k}\left\Vert\left(\dfrac{\mathbf{x}(d,:)\mathbf{x}(d,:)^T}{\left\Vert\mathbf{x}(d,:)\right\Vert^2} - I\right)\mathbf{c}'\right\Vert^2\\
    &= \underset{\mathbf{c}'}{\argmin}\frac{1}{\left\Vert \mathbf{c}' \right\Vert^2 }\mathbf{c}'^{T}\sum_{\mathbf{x} \in S_k}\left(I-\dfrac{\mathbf{x}(d,:)\mathbf{x}(d,:)^T}{\left\Vert\mathbf{x}(d,:)\right\Vert^2}\right)\mathbf{c}'\;\;,
\end{align*}

and by replacing $\sum_{\mathbf{x} \in S_k}(I-\dfrac{\mathbf{x}(d,:)\mathbf{x}(d,:)^T}{\left\Vert\mathbf{x}(d,:)\right\Vert^2})$ with $M$, we obtain the following minimization problem

\begin{align*}
\mathbf{c}_k = \underset{\mathbf{c'}}{\argmin} \frac{\mathbf{c}'^{T}M\mathbf{c}'}{\left\Vert\mathbf{c}'\right\Vert^2}\;\;,
\end{align*}

which achieves its minimum value with the eigenvector of $M$ corresponding to the smallest eigenvalue. We refer further interested readers to the properties of the Rayleigh quotient with parameters $\mathbf{c}'$ and $M$~\cite{golub}. 

After successfully constructing the two building blocks of kSC for our multivariate case, we present the pseudo code of our iterative algorithm \textit{multivariate kSC} (m-kSC), \textit{cf}. Alg.~\ref{alg:m-kSC}.

\begin{algorithm}
\caption{m-kSC Algorithm}
\label{alg:m-kSC}
\begin{algorithmic}[1]
\Statex\textbf{Input}:$\{\mathcal{X},K\}$ where $\mathcal{X} \in \mathbb{R}^{N\times D\times M}$ is the tensor containing $N$ multivariate time series and $K$ is number of clusters.
\Statex\textbf{Output}: $\{\mathcal{C},S\}$ where $\mathcal{C} \in \mathbb{R}^{K\times D\times M}$ is the tensor of cluster centroids and $S$ contains each cluster assignments.
\State Initialize cluster assignments $S$ randomly

\While{$S$ changes on every iteration}
\For{$k = 1:K$}
\For{$d = 1:D$}
\State $M = \sum_{\mathbf{x}_n \in S_k}(I-\dfrac{\mathbf{x}_n(d,:)\mathbf{x}_n(d,:)^T}{\left\Vert\mathbf{x}_n(d,:)\right\Vert^2})$
\State $\mathbf{C}(k,d,:) = $ Smallest eigenvector of $M$.
\EndFor
\EndFor
\For{$n=1:N$}
\State $k = \underset{k=1,...,K}{\argmin}dist(\mathbf{c}_k,\mathbf{x}_n)$ using Eq.~\ref{eq:distance}
\State $S(n) = k$
\EndFor
\EndWhile
\end{algorithmic}
\end{algorithm}

\subsubsection{Finding the optimal K}

As with many kMeans based approaches, m-kSC also needs the number of clusters as a parameter. A common practice to find the optimal $K$ involves post-processing, where we run the clustering algorithm several times for different $k$ values. In this case, a trade off between the final loss function value and model complexity decides the optimal $K$. However, to avoid re-running the algorithm for every possible $k$, here we adopt \textit{dip-dist}, the iterative strategy proposed by~\citet{dipmeans}. We choose the dip-dist technique over others~\cite{xmeans,gmeans} as the only assumption dip-dist makes is the unimodality of pairwise distances of cluster members. It suggests that an optimal cluster structure should not have more than a single mode among pairwise distances of its members (multimodality). The density of pairwise distance values should reach to maximum around the mode and decay while moving away.

Multimodality of each cluster is checked by its members' distance to each other. First, we calculate the distance of each time series to the other members in the cluster. Then, we sort the distance vector in decreasing order. If the null hypothesis of unimodality in the distribution of the sorted distances is rejected by Hartigan's dip test \cite{hartigan_dip} with a significance level $p$-value$<\alpha$, it is considered to be a splitter in the cluster. Finally, if the number of splitters is higher than a given threshold $v$, we split the cluster into two disjoint sub-clusters. Our splitting strategy involves a local search with m-kSC where k is set to 2. Heuristically, we choose the least squared error objective function centroids over 10 runs. We iteratively split clusters until there is no clusters left with ratio of splitters greater than the given threshold. As suggested by~\citet{dipmeans}, at each iteration, we only split the cluster with maximum number of splitters. We report our complete dip-test based multidimensional spectral centroid algorithm dipm-SC, \textit{cf}. Alg.~\ref{alg:dipm-SC}. 
We refer further interested readers to~\citet{dipmeans} for kMeans extension and to~\citet{hartigan_dip} for the Hartigan's dip test. We make the source code and the experimental settings available at \url{www.public.asu.edu/~mozer/dipm-SC/dipm_source_code.zip}.

The computational cost of m-kSC is dominated by the calculation of matrix M for each $x_n$ and for d dimensions ($\mathcal{O}(m^2dn)$) and the eigendecomposition of M for k clusters ($\mathcal{O}(m^3dk)$). Multimodality test in dipm-SC for each cluster k has a complexity of $\mathcal{O}(k(bn\log n+n^2))$. So, the total complexity of the dipm-SC becomes the maximum of the $\{\mathcal{O}(m^2dn),\mathcal{O}(m^3dk),\mathcal{O}(k(bn\log n+n^2))\}$.

\begin{algorithm}[t!]
\caption{dipm-SC Algorithm}
\label{alg:dipm-SC}
\begin{algorithmic}[1]
\Statex\textbf{Input}:$\{\mathcal{X},\alpha, s\}$ where $\alpha$ is the significance level, and $v$ is the split threshold.
\Statex\textbf{Output}: $\{\mathcal{C}^{*},S^{*}\}$.
\State Assign time series to one cluster $S = ones(N,1)$
\For{$i=1:N$}
    \For{$j=1:N$}
		\State $\mathbf{d}(i,j) = dist(\mathcal{X}(i,:,:),\mathcal{X}(j,:,:))$ Equation \ref{eq:distance}
	\EndFor
\EndFor

\While{$max(score_k)>v$}
	\For{$k=1:K$}	
	\State $score_k = checkClusterModality(\mathbf{d}(S_k,S_k),S_k,\alpha)$
	\EndFor
	\If{$max(score_k) > v$}
	\State$k = \underset{k}{\argmax}(score_{k})$
	\State$[\mathbf{c}_1,\mathbf{c}_2] = $splitCluster$(\mathcal{X}(S_{k},:,:))$
	\State$\mathcal{C}(k,:,:) = \mathbf{c}_1$
	\State$\mathcal{C}(K+1,:,:) = \mathbf{c}_2$
	\State$K = K+1$
	\State$[\mathcal{C},S] = $m-kSC$(\mathcal{X},K,\mathcal{C})$
	\EndIf
\EndWhile
\end{algorithmic}
\end{algorithm}

\section{Evaluation}

We evaluate the results of our algorithm applied to one synthetically generated and two real-world multivariate time series datasets: GitHub and Twitter. We refer interested readers to \url{https://www.public.asu.edu/~mozer/dipm-SC/supp\_material.pdf} for the comparison of the dipm-SC with extension of other shape-based time series clustering algorithms on synthetically generated time series dataset and the GitHub analysis. For the sake of brevity of this paper, here we present our experimental results for Twitter hashtag dataset. First, we perform the qualitative analysis by studying different behavioral patterns of the detected multivariate centroids. Secondly, we characterize the clusters based on their "periodicity", locating them on a scale from periodic to viral. We expect that periodic time series will behave steadily and that the viral time series will exhibit significant spikes over short time periods. To this aim, we use two previously introduced metrics; \textit{burstiness} and \textit{memory}~\cite{goh2008burstiness}. These two metrics are computed from the distribution of the inter-event time $P(\tau)$ between consecutive events of time series data. In particular, given the inter-event time distribution $P(\tau)$ having mean $m_{\tau}$ and standard deviation $\sigma_{\tau}$, we calculate burstiness as

\begin{equation}
B = \frac{\sigma_{\tau}-m_{\tau}}{\sigma_{\tau}+m_{\tau}}\;\,
\label{eq:burstiness}
\end{equation}
where $B \in[-1,1]$: $B = 1$ corresponds to the most bursty sequence of events, and $B=-1$ is a completely regular (periodic) sequence.

This measure however is not enough to characterize the correlations taking place between consecutive events. Correlation of consecutive events can be thought as a memory process. A measure that can be used to compute such memory coefficient between consecutive events $(\tau_i, \tau_{i+lag})$ is defined as

\begin{equation}
M = \frac{1}{n_{\tau}-1}\sum^{n_{\tau}-1}_{i=1}\frac{\left(\tau_i-m_1\right)\left(\tau_{i+lag}-m_2\right)}{\sigma_{1}\sigma_{2}}\;\,
\label{eq:memory}
\end{equation}
where, $n_{\tau}$ is the number of intervals between events in the time series, while $m_{1,2}$ and $\sigma_{1,2}$ are respectively the sample mean and standard deviation of $\tau_{i,i+lag}$'s. We set the value of lag based on the empirically observed value of periodicity for each case.

\section{Datasets description}
We collected tweets starting from February 14th until March 6th, 2018 related to the Parkland school shooting event from GNIP. To this aim, we provided 140 terms (hashtags, words, phrases) related to the event and the broader gun debate in the United States.\footnote{\url{https://www.public.asu.edu/~mozer/dipm-SC/GNIP\_query\_list.txt}} GNIP  provides  any tweet that contains at least one of the queried terms, while not suffering from any bias implications of free API streams~\cite{fred}. Our data contains $~23.7$M tweets coming from $~3.7$M users. 

We then identify the top $1,000$ most occurring (popular) hashtags in our dataset and build hourly time series for each of them. Alongside with hourly counts of each hashtag, we construct the hourly ratio of tweets sent by bot accounts to the total hourly volume. To identify the most obvious bot accounts, we used the free API provided by \textit{Botometer}~\cite{botometer}. We  tested about $5\%$ of the most active accounts ($193$K approximately) and found that $~19$K of them are likely bots---in line with other research on bot frequency~\cite{varol}. 

As the third and fourth dimensions, we consider the hourly rate of positive and negative sentiment tweets. We use an off-the-shelf short text classification tool SentiStrength~\cite{sentistrength} to detect positive and negative sentiment of each tweet where hashtags occur. SentiStrength provides scores for positive and negative sentiments separately. We aggregate each sentiment's score on an hourly basis by simply summing them. Finally, we build the hourly sentiment ratio time series for each hashtag by normalizing the sentiment volumes by total tweet volumes. 

Therefore, the timeline of each hashtag over time is represented by 4 separate dimensions: total count, bot tweet rate, positive sentiment rate, and negative sentiment rate.
\vspace{-5mm}
\section{Popularity Gain Patterns on Twitter}
\begin{figure}[t!]
    \centering
    \subfloat[Cluster-P]{
    \includegraphics[width=0.33\linewidth]{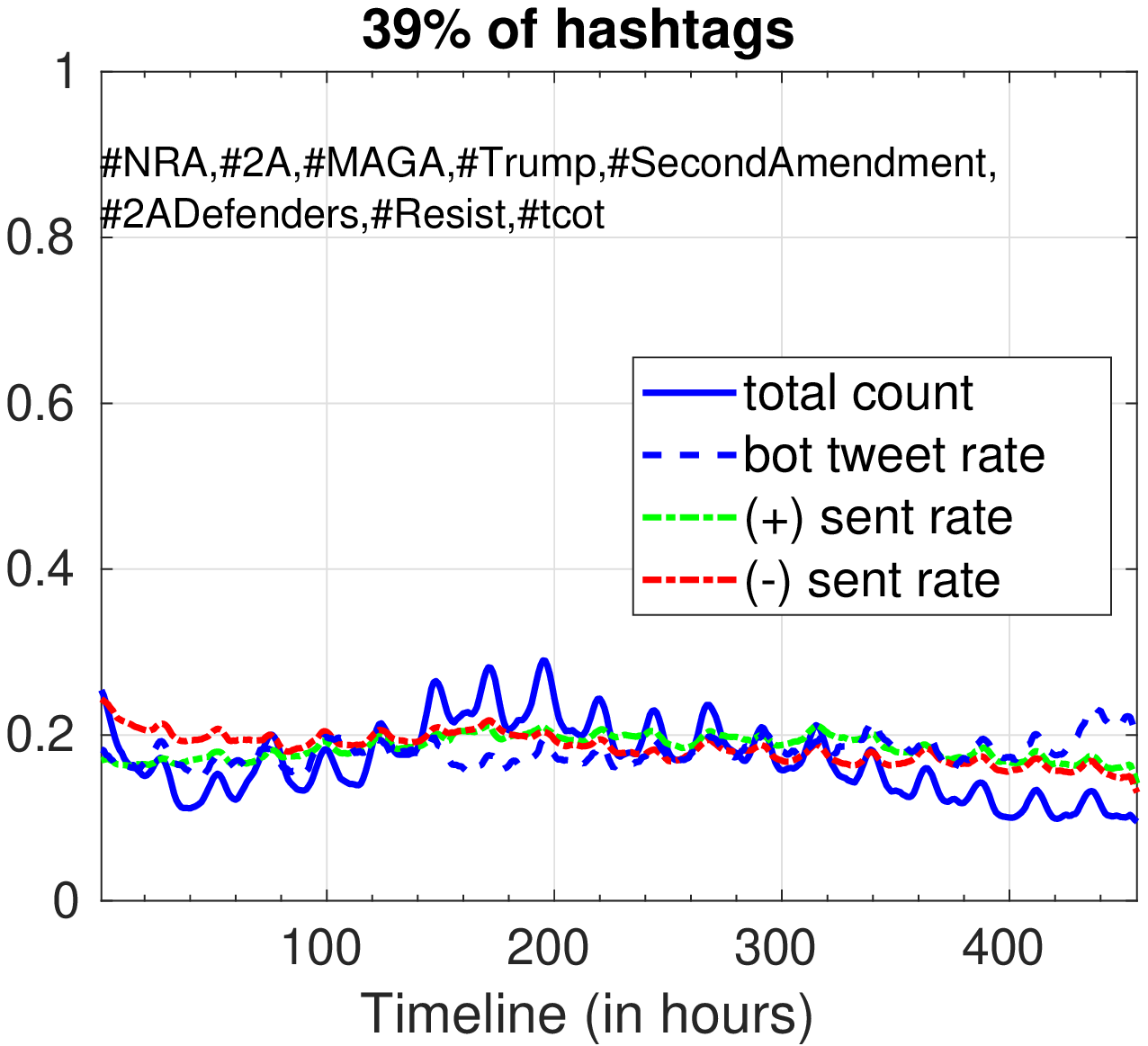}
    }
    \subfloat[Cluster-B1]{
    \includegraphics[width=0.33\linewidth]{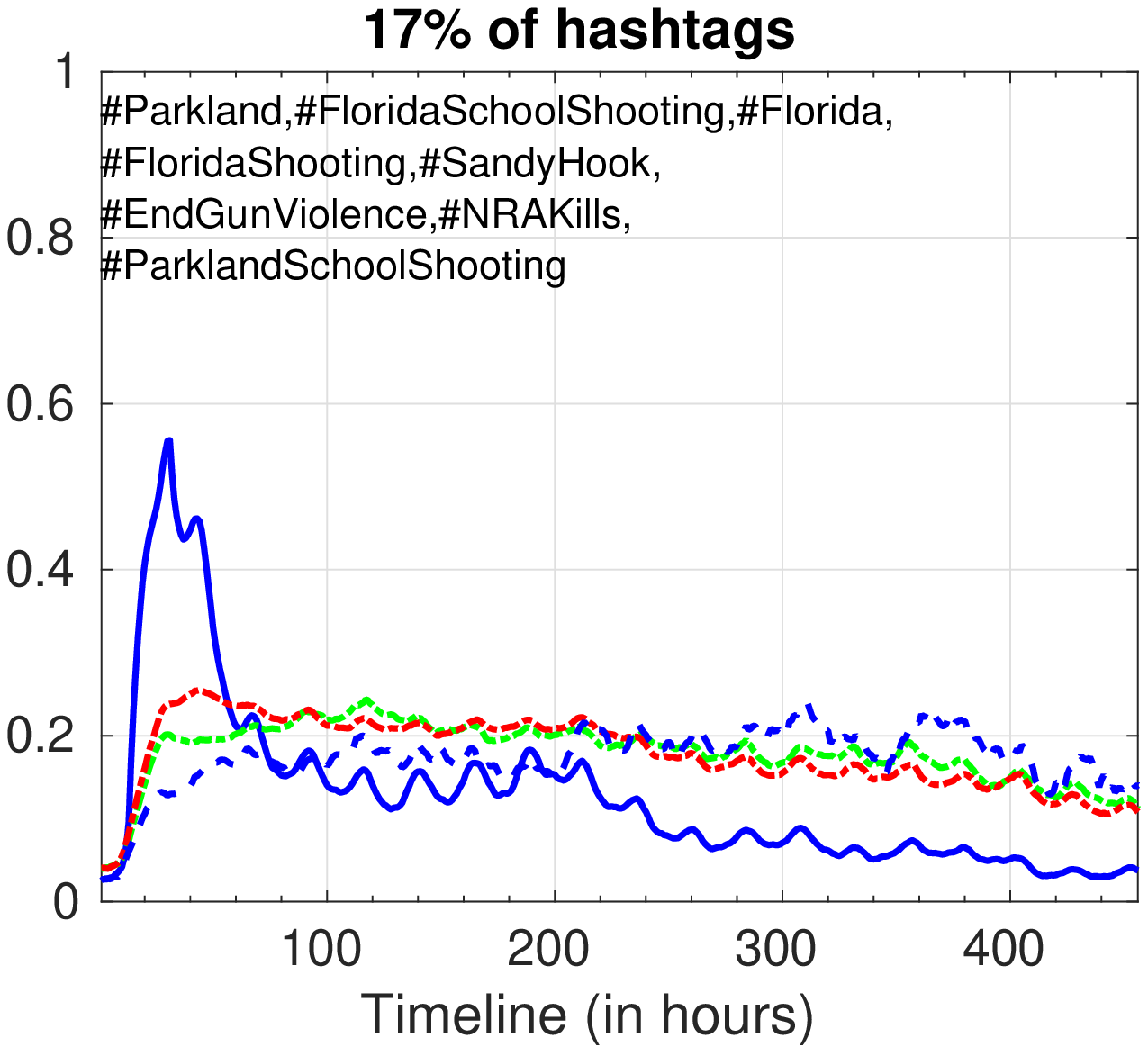}
    }
    \subfloat[Cluster-B2]{
    \includegraphics[width=0.33\linewidth]{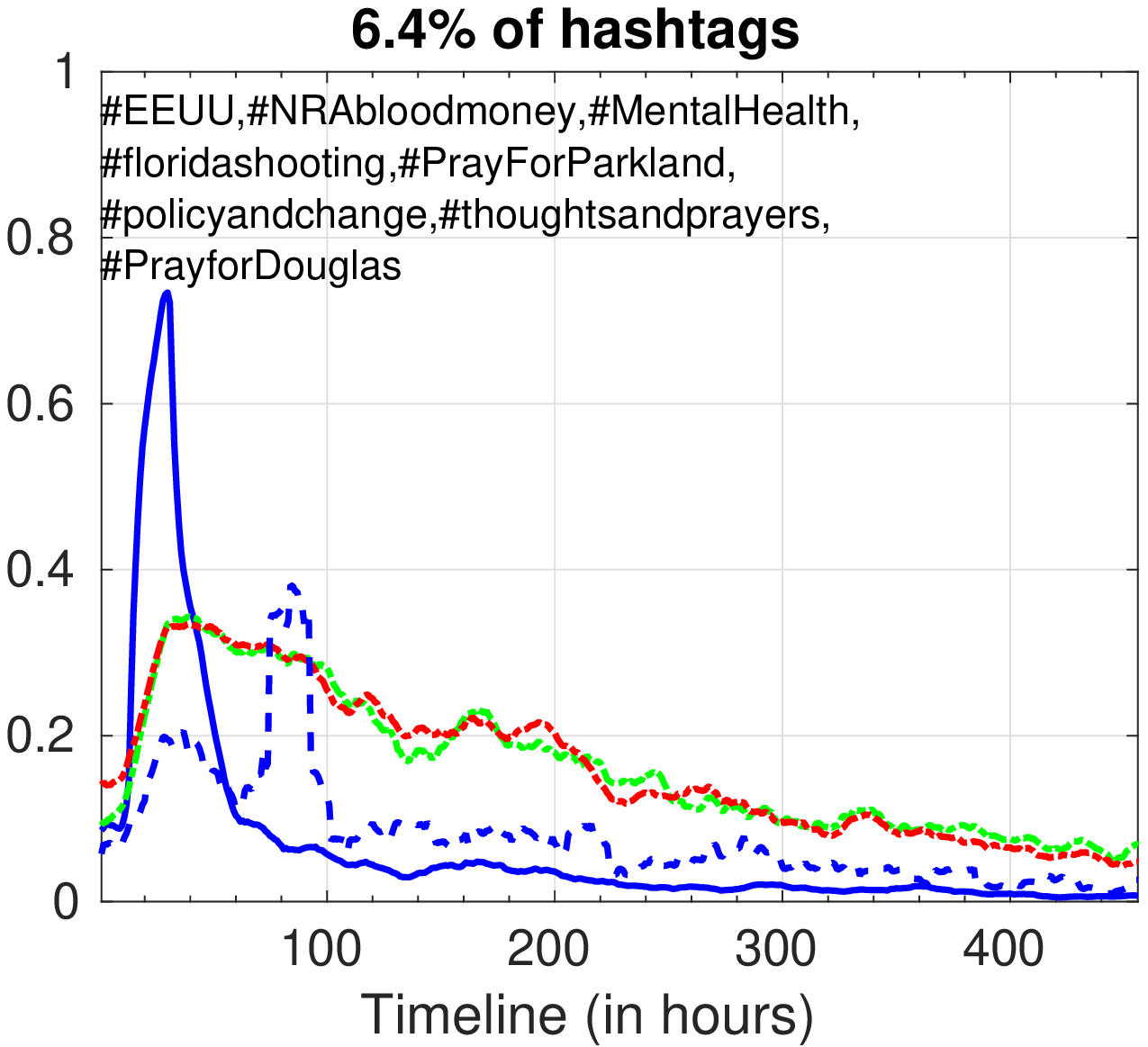}
    }\\
    \subfloat[Cluster-B3]{
    \includegraphics[width=0.33\linewidth]{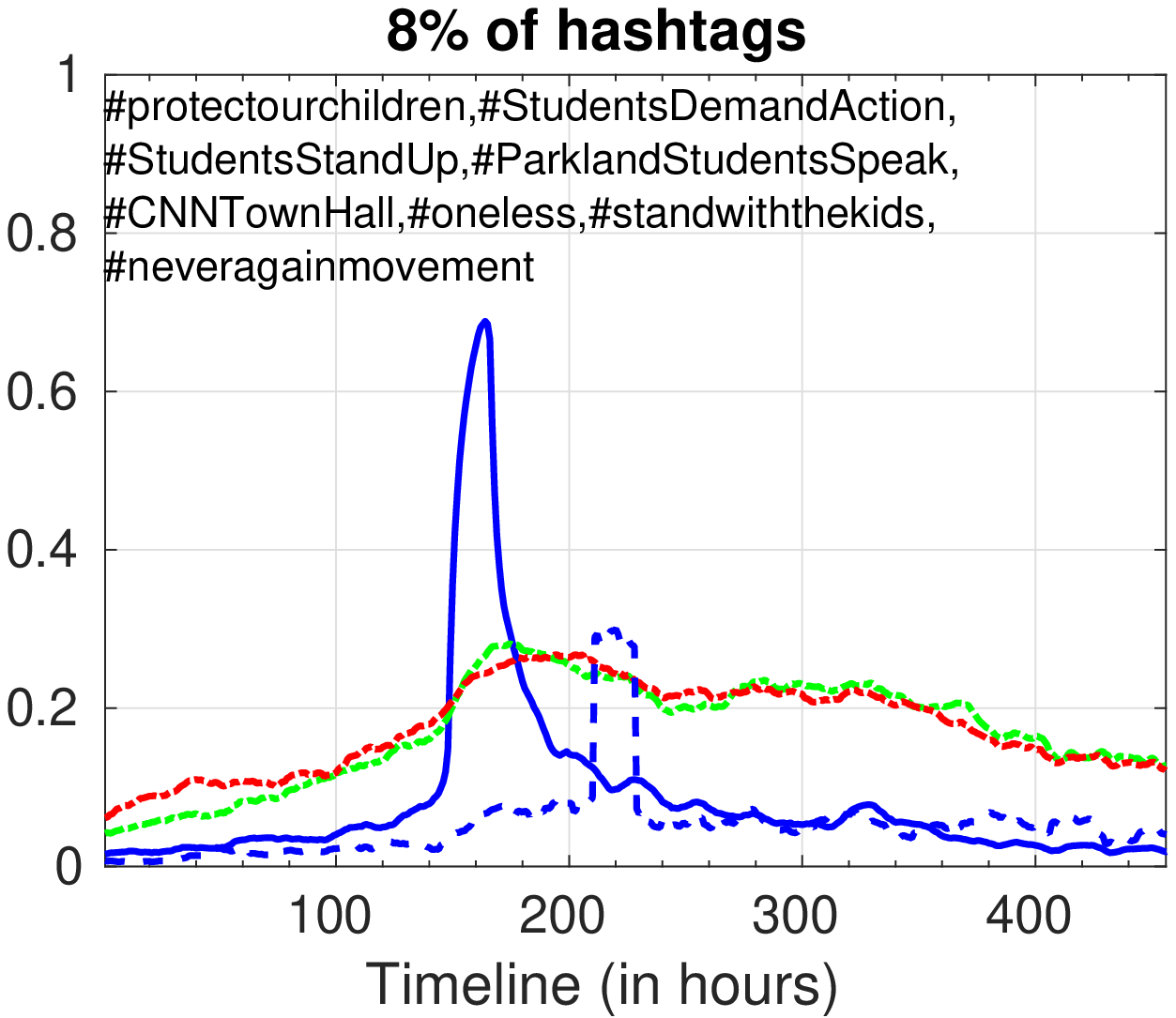}
    }
    \subfloat[Cluster-Boycott1]{
    \includegraphics[width=0.33\linewidth]{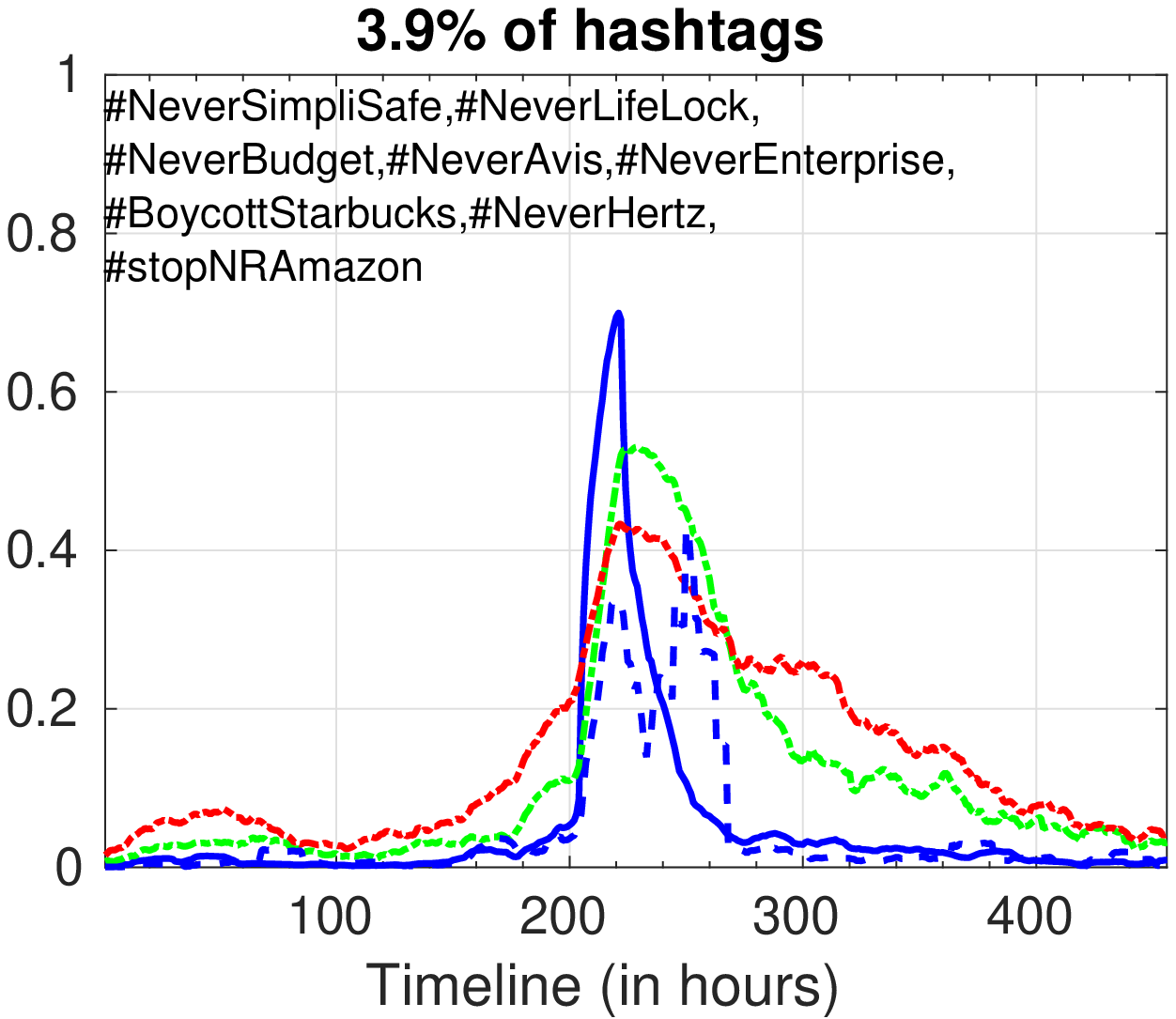}
    }
    \subfloat[Cluster-Boycott2]{
    \includegraphics[width=0.33\linewidth]{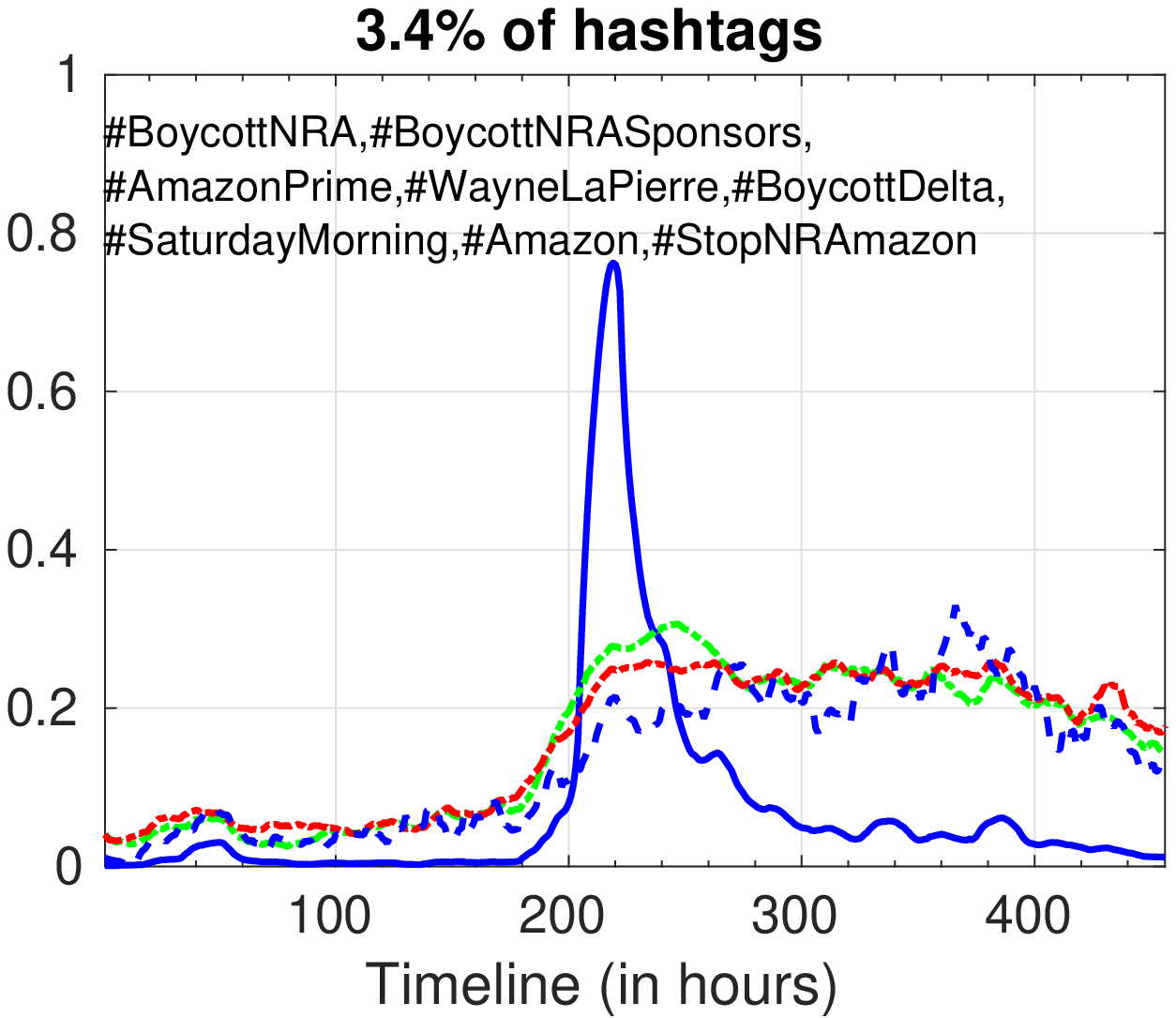}
    }
\caption{Shape of the uncovered clusters of Twitter}
\label{fig:Twitter-Shapes}
\end{figure}

In this section we present Twitter hashtag timeline analysis with the help of multidimensional shape clusters identified by our dipm-SC algorithm. We use 24 hours as our shifting parameter. We also apply gaussian smoothing of window size 24 to eliminate noise. In total we are able to identify 13 clusters in different timeline shapes. Here we present 6 clusters which spans nearly 78\% percent of popular hashtags under study. We leave other 7 smaller clusters reachable externally for brevity of the paper (\url{http://www.public.asu.edu/~mozer/dipm-SC}). We report these 6 most popular timeline shapes in Fig.~\ref{fig:Twitter-Shapes}.

First, we observe that cluster-P in Figure \ref{fig:Twitter-Shapes}a is significantly different in shape when compared with others. In particular, it does not show any burst of activity in each of the four dimensions considered (popularity, sentiment ratios, and bot tweet ratio). On the contrary, this cluster shows a steady pattern with daily periodic activities. 

On the other hand, the remaining 5 clusters are characterized by a sudden spike in the popularity signal at a certain point in time and different trends among the other dimensions. The popularity burst of Cluster-B1 and B2 occurs at the same time. Spike in popularity gain considerably differs in the tail part. Cluster-B1 shows a longer tail while Cluster-B2 dies off sooner. Cluster-B2 temporal pattern shows not only a decreasing trend in popularity but also a faster decay in both sentiment and bot ratios then the one we observe in cluster-B1. We notice an analogous discrepancy among the last two clusters (i.e., Boycott1 and Boycott2).

Finally, we analyze the temporal coherence of hashtags clustered together. We note that our algorithm does not consider semantics while clustering timeline of hashtags. The detected clusters are indeed identified because they share similar temporal patterns among different dimensions. We report the top 8 closest hashtags to the cluster centroids in the top part of each panel in Figure \ref{fig:Twitter-Shapes}.

Cluster-P contains broader issue related hashtags such as \#NRA, \#2A(second amendment), \#MAGA, \#Trump, which show steady and periodic behavior among dimensions. Cluster B1 and B2 exhibits spikes around the time when the tragic shooting takes place. In cluster-B1 we observe event-related long lasting hashtags, while in cluster-B2 we observe event-related expiring hashtags such as \#prayfor variants, \#MentalHealth. Cluster B3 is unique in terms of its spike time when student survivors start the \#neveragain movement and appear in a town hall organized by CNN. Lastly, we notice two boycott related hashtags dominated clusters spiking in popularity gain around the same time. It overlaps with the time of students' call to boycott NRA and companies that shares financial interests with it. It is easy to see these two clusters differing in their tail parts. Cluster-Boycott2 (longer popularity gain tail) shows more NRA \& Amazon related hashtags while Cluster-Boycott1 involves other companies names. We will later present statistically significant difference in cumulative popularity gains between these two clusters.

\subsection{Interplay of Dimensions}
The key piece of information we acquire through this analysis is how popularity spike of a hashtag correlates with increased sentiment ratios and bot involvement. For every bursty popularity gain, we observe an increase in all other dimensions. Previous studies on bursty attention mechanisms of online content usually focus on the characteristic of the fat-tail \citep{yang2011patterns,lehmann2012dynamical,ratkiewicz2010characterizing}. Here, we observe that both sentiment and bot tweet ratios stay more steady with a longer tail of popularity (e.g., B1, B3,  and Boycott2) than dying ones (e.g., B2, Boycott1). We evidence it by fitting a linear curve and measuring the average slope of cluster members. When compared, hashtags belonging to B1 have a higher slope than the ones in B2 in sentiment and bot involvement dimensions after the spike. We observe the same pattern between Boycott2 and Boycott1 clusters, where Boycott2 has a higher slope in all three dimensions after the spike.

Another interesting behavior we direct attention to is the sudden increase in bot tweet ratio after the popularity spike in cluster B2, B3 and Boycott1. These clusters are characterized by a weaker tail in popularity gain after a burst in popularity. This signals a steady bot involvement for a while although abandonment of the overall activity takes place.

\begin{figure}[hbt!]
\centering
\subfloat[]{
    \includegraphics[width=0.46\linewidth]{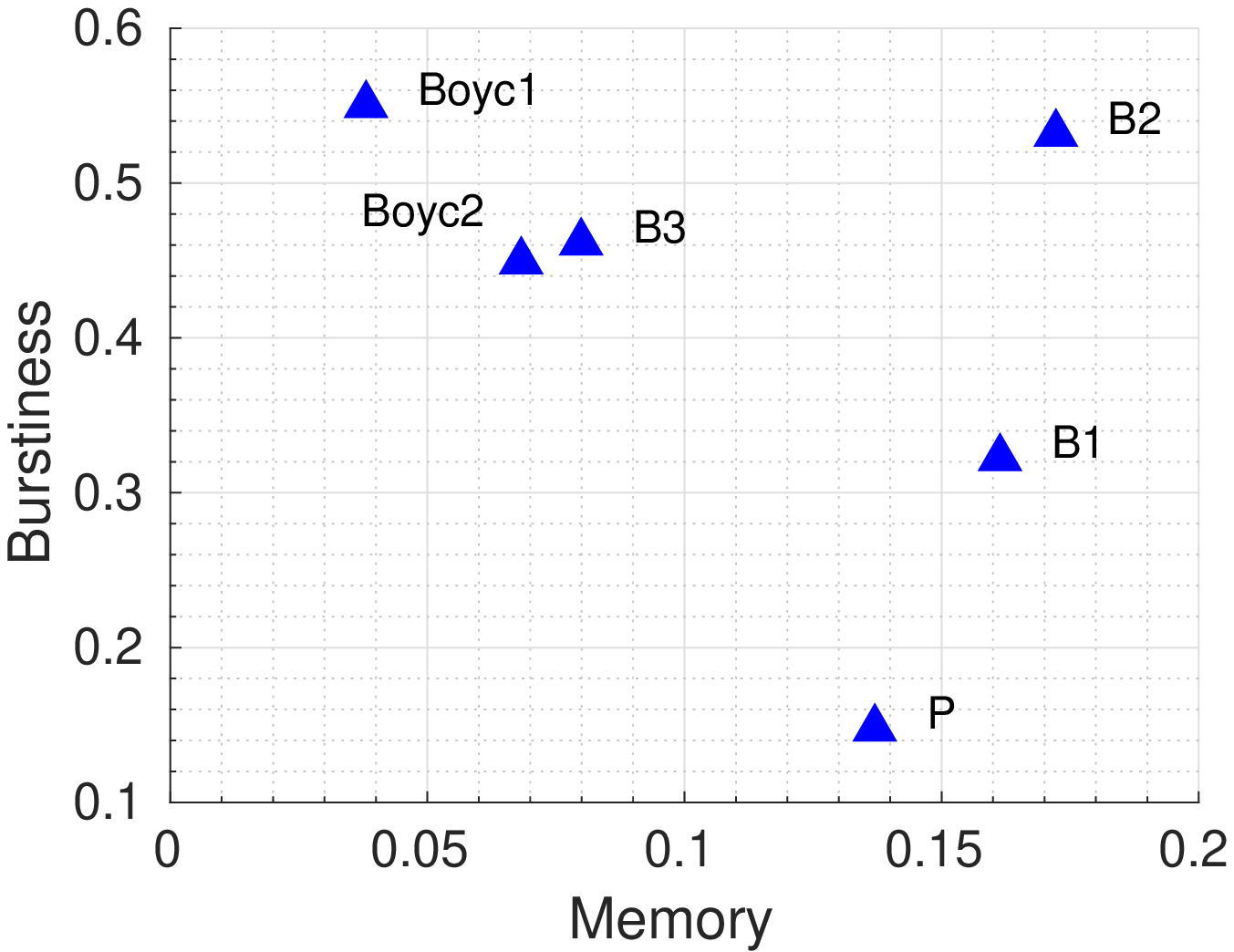}
    \label{fig:twitter_burst_mem}
    }\quad
\subfloat[]{
    \includegraphics[width=0.46\linewidth]{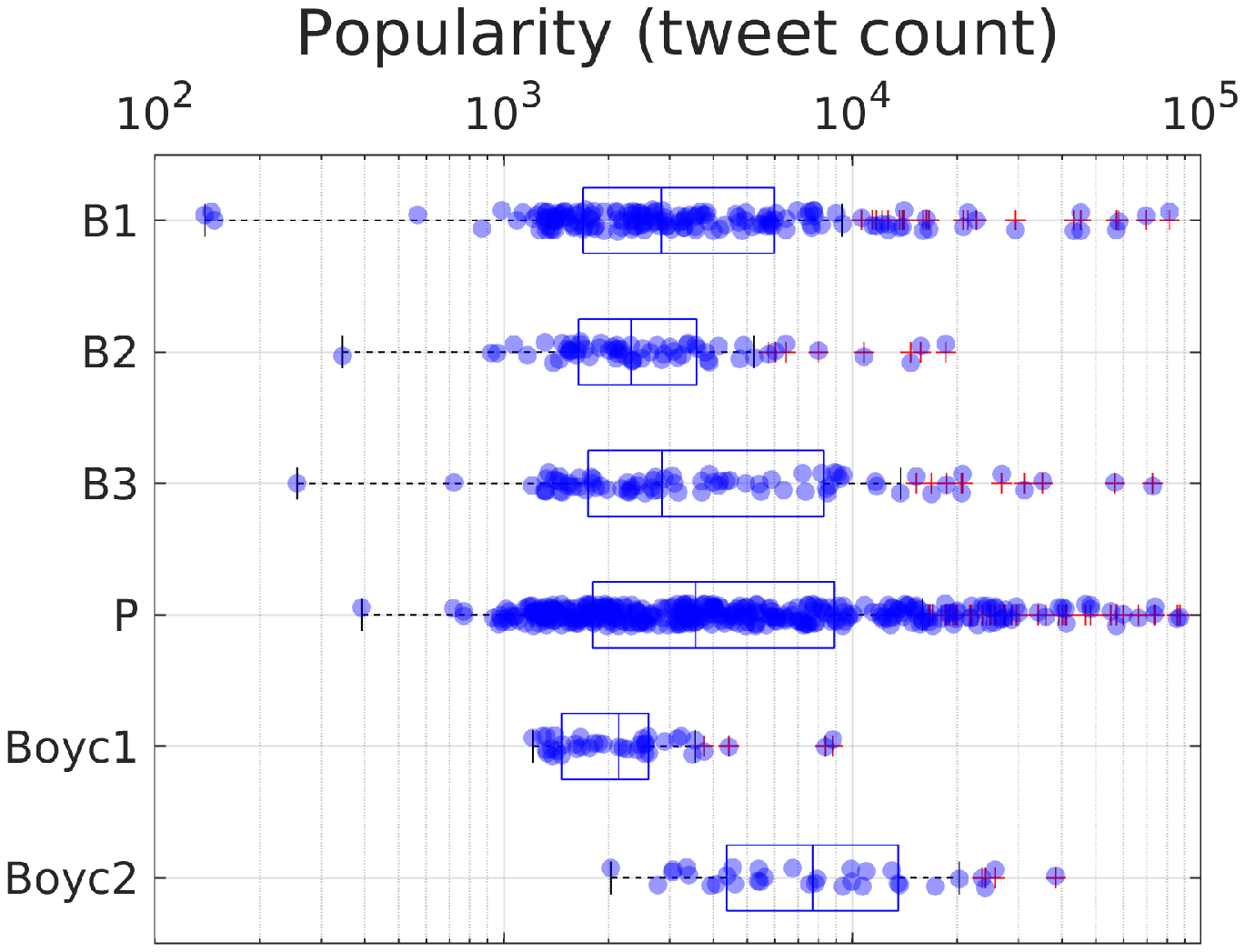}
    \label{fig:Twitter-Popularity}
}
\caption{Burstiness and Memory of Clusters. Figure (a): Burstiness and memory of Twitter clusters. Figure (b): Cumulative popularity of cluster members at the end of the data timeline. Each blue dot represents a hashtag.}
\label{fig:twitter_cluster_stats}
\end{figure}

\subsection{Burstiness and Memory of Clusters}
First, we quantify popularity gain's temporal behavior with the described burstiness and memory metrics. We present average burstiness and memory of each cluster in  Figure \ref{fig:twitter_burst_mem}. As expected, cluster-P has the lowest burstiness while enjoying higher values of memory. Lowest memory and highest burstiness belongs to Boycott1 cluster.

Next, we investigate if different multi-dimensional timeline shapes lead to significant difference in hashtags' cumulative popularity gain at the end of the time window we observe. Since this analysis is based on the timeline of a hashtag rather than a lifespan, we do not align hashtags in their creation date. That is why, for example, comparing Cluster-B1 against Cluster-B3 is not a meaningful test. However, it is noteworthy that two-sample Kolmogorov-Smirtnov test rejects the null hypothesis for Cluster-Boycott1 and Cluster-Boycott2 of which we know the beginning of campaigns. Differing distributions for these two shape clusters can be also observed from Figure \ref{fig:Twitter-Popularity}. Moreover, it is also appealing to see no cumulative popularity distribution difference between hashtags of steady cluster P and bursty cluster B1 (p-value 0.056).

\vspace{-2mm}

\section{Related Work}
Our work overlaps with the current literature among two branches. First is the task of clustering time series, and the second is the task of uncovering temporal patterns of online content.

Clustering time series has been a challenging task and research area that produced vast amount of literature~\cite{Aghabozorgi2015TimeseriesC}. Usually, when external feature learning or modeling~\cite{modeling,deep,model_cluster} is not imposed, clustering of time series involves choosing a proper time series distance function and an algorithm. Numerous works have introduced various distance measures to calculate proximity of time series data~\cite{survey1, survey2}. There have been partitional~\cite{kshape, ksc, kdba, kavged}, subsequence matching~\cite{subsequence,subsequence_useless}, hierarchical~\cite{hierarchical} approaches using these different time series distance measures. Moreover, a major body of work~\cite{ushapelet, ushapelet2} exists in subsequence matching based time series clustering where they identify shorter most identifying portions of time series data also known as shapelets to group them. For the multivariate time series data, same categorizations can be made as modeling based~\cite{toeplitz, multi_hmm}, and variants of generalizing univariate solutions to multivariate cases~\cite{multi-dtw, multi-pca}. Our approach falls into the second category where we extend existing distance functions available for univariate time series data and update centroid finding (i.e. averaging function) accordingly.

As the second line of work from literature, we compare our effort with analysis of temporal patterns of online content. There is significant amount of literature focusing on characterizing and modeling bursts and long tail dynamics~\cite{barabasi_tail, spikem, lehmann2012dynamical}. First and foremost, our study does not focus on modeling, rather it finds clusters of shapes inherent to online temporal behavior. However, we show that our cluster shapes can be identified distinctively by their burstiness and periodicity through postprocessing. Our work also differentiates itself from others~\cite{blogshapes,spikem,ksc} by analyzing lifetime or a longer timeline of online entities rather than focusing on a window where activity peaks occur and omitting the rest.
\vspace{-2mm}
\vspace{-3mm}
\section{Conclusion}
\vspace{-2mm}
In the present work, we set to study the temporal patterns that lead a user's content to gain popularity in online platforms. In particular, our study tackled the following problems: (i) uncovering the different temporal patterns of popularity in online platforms; (ii) studying how different dimensions' interplay; and (iii) proposing dipm-SC, a novel algorithm that extends the state-of-the-art models allowing to cluster multidimensional time series of events at a time.

First, we compared our method with other extensions of univariate time series clustering algorithms on synthetic data, where we successfully demonstrated its higher accuracy. We then applied our framework on the multivariate time series data deriving from two major online platforms: GitHub and Twitter. Through these applications, we showcased the efficacy of our method in uncovering fundamental pattern of popularity in online contexts. Our method can indeed be used not only to find the overall difference between time series shapes, such as bursty versus steady behaviors, but it can also uncover differences in these two main trends depending on the time in which popularity is acquired. Moreover, the results provided by our approach are easily interpretable and sheds light on analyzing the interplay of multiple dimensions. In the Twitter scenario, we found that the uncovered clusters are temporally coherent in the hashtags used and related to certain types of events, such as the Parkland school shooting and boycott campaigns.

In conclusion, we devised a methodology that extend the state-of-the-art literature in the area of multivariate time series clustering and uses a similarity metric based on the shape of the time series. Moreover, our method can be used both to uncover fundamental patterns based solely on the shapes of the time series and on the temporal occurrence of the events. Our approach also provide a way of studying popularity of online content and understanding their dynamics over time.

\subsubsection{Acknowledgements}
The authors are grateful to the Defense Advanced Research Projects Agency  (DARPA), contract W911NF-17-C-0094, for their support. 

\bibliographystyle{splncsnat}
\bibliography{main}
\end{document}